\documentclass[10pt,twocolumn,letterpaper]{article}
\usepackage{booktabs}
\usepackage{iccv}
\usepackage{times}
\usepackage{epsfig}
\usepackage{amsmath}
\usepackage{amssymb}
\usepackage{adjustbox}
\usepackage{subcaption}
\usepackage{bm}
\usepackage{graphicx}
\usepackage{caption}
\usepackage{float}
\usepackage{nth}

\usepackage{color}

\usepackage[breaklinks=true,bookmarks=false]{hyperref}

\newcommand\real{\mathbb{R}}
\newcommand{\pf}[2]{{\color{red}#1}}
\newcommand{\hamed}[1]{#1}

\newcommand{\hamedcom}[1]{}

\iccvfinalcopy 

\def\iccvPaperID{****} 

\ificcvfinal\fi
\begin{document}

\title{Representation Learning by Learning to Count}


\author{Mehdi Noroozi$^1\quad$ Hamed Pirsiavash$^2 \quad$  Paolo Favaro$^1$\\
\hspace{0em} University of Bern$^1\quad   $ University of Maryland, Baltimore County$^2$\\
{\hspace{-0em} $\texttt{\small \{noroozi,favaro\}@inf.unibe.ch}   \quad  \texttt{\small \{hpirsiav@umbc.edu\}}$}
}

%
%
%
\maketitle
\thispagestyle{empty}

\begin{abstract}
We introduce a novel method for representation learning that uses an artificial supervision signal based on counting visual primitives. This supervision signal is obtained from an equivariance relation, which does not require any manual annotation. We relate transformations of images to transformations of the representations. More specifically, we look for the representation that satisfies such relation rather than the transformations that match a given representation. In this paper, we use two image transformations in the context of counting:
scaling and tiling.
The first transformation exploits the fact that the number of visual primitives should be invariant to scale. The second transformation allows us to equate the total number of visual primitives in each tile to that in the whole image.
These two transformations are combined in one constraint and used to train a neural network with a contrastive loss. \hamedcom{is this loss important here?}
The proposed task produces representations that perform on par or exceed the state of the art in transfer learning benchmarks.
\end{abstract}

\section{Introduction}



We are interested in learning representations (features) that are discriminative for semantic image understanding tasks such as classification, detection, and segmentation. A common approach to obtain such features is to use supervised learning. However, this requires manual annotation of images, which is costly, time-consuming, and prone to errors.
In contrast, unsupervised or self-supervised feature learning methods exploiting unlabeled data can be much more scalable and flexible.

Some recent feature learning methods, in the so-called \emph{self-supervised learning} paradigm, have managed to avoid annotation by defining a task which provides a \emph{supervision signal}. For example, some methods recover color from gray scale images and vice versa \cite{colorful,larsson2016learning,splitBrain, larsson2017colorproxy}, recover a whole patch from the surrounding pixels \cite{ContextEncoder}, or recover the relative location of patches \cite{context,noroozi2016}. These methods use information already present in the data as supervision signal so that supervised learning tools can be used. A rationale behind self-supervised learning is that pretext tasks that relate the most to the final problems (\eg, classification and detection) will be more likely to build relevant representations.

\begin{figure}[t]
\begin{center}
\includegraphics[width=.79\linewidth,trim={1.6cm 5cm 2cm 3cm},clip]{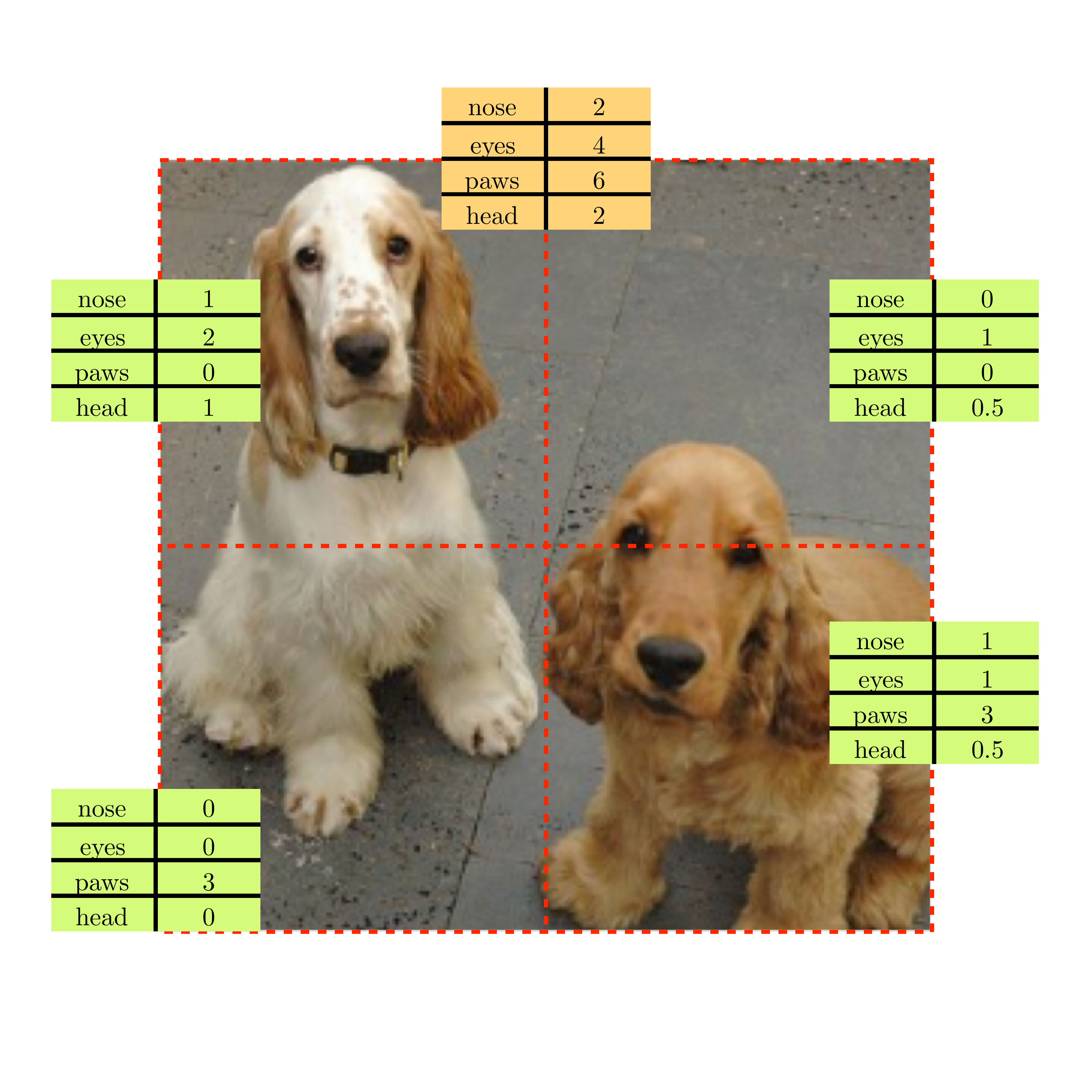}
\end{center}
   \caption{The number of visual primitives in the whole image should match the sum of the number of visual primitives in each tile (dashed red boxes).}
\label{fig:learningCounting}
\end{figure}

As a \hamed{novel} candidate pretext task, we propose \emph{counting visual primitives}. It requires discriminative features, which can be useful to classification, and it can be formulated via detection. 
To obtain a supervision signal useful to learn to count, we exploit the following property: If we partition an image into non-overlapping regions, the number of visual primitives in each region should sum up to the number of primitives in the original image (see the example in Fig.~\ref{fig:learningCounting}). 
We make the hypothesis that the model needs to disentangle the image into high-level factors of variation, such that the complex relation between the original image and its regions is translated to a simple arithmetic operation~\cite{BetterMixing,DGAN16}. Our experimental results validate this hypothesis both qualitatively and quantitatively. 

While in this work we focus on a specific combination of transformations, one can consider more general relationships (\ie, beyond counting, scaling, and tiling) as supervision signals. The same procedure that we introduce is therefore applicable to a broader range of tasks as long as it is possible to express the transformation in feature space caused by a transformation in image space \cite{LencV15}.

Our contributions are:
1) We introduce a novel method to learn representations from data without manual annotation; 2) We propose exploiting counting as a pretext task and demonstrate its relation to counting visual primitives; 3) We show that the proposed methodology learns representations that perform on par or exceed the state of the art in standard transfer learning benchmarks.

\section{Prior Work}

In this work we propose to learn a representation without relying on annotation, a problem that is typically addressed via unsupervised learning. An example of this approach is the autoencoder \cite{AE06,DAE06}, which reconstructs data by mapping it to a low-dimensional feature vector. A recent alternative approach is \emph{self-supervised learning}, which is a technique that substitutes the labels for a task with \emph{artificial} or \emph{surrogate} ones. In our work such artificial labels are provided by a counting constraint. 
In many instances, this technique can be seen as recasting the unsupervised learning problem of finding $p(\mathbf{x}) = p(\mathbf{x}_1, \mathbf{x}_2)$, where $\mathbf{x}^\top = [\mathbf{x}_1^\top~\mathbf{x}_2^\top]$ is a random variable, as a partly supervised one of finding $p(\mathbf{x}_2|\mathbf{x}_1)$, so that we can write $p(\mathbf{x}_1,\mathbf{x}_2) = p(\mathbf{x}_2|\mathbf{x}_1)p(\mathbf{x}_1)$ (cf. eq.~(5.1) in \cite{deeplearningbook}). 
In our context, the data sample $\mathbf{x}$ collects all available information, which can be just an image, but might also include egomotion measurements, sound, and so on. In the literature, self-supervised methods do not recover a model for the probability function $p(\mathbf{x}_1)$, since $p(\mathbf{x}_2|\mathbf{x}_1)$ is sufficient to obtain a representation of $\mathbf{x}$.
Most methods are then organized based on the choice of $\mathbf{x}_1$ and $\mathbf{x}_2$, where $\mathbf{x}_2$ defines the surrogate labels.
Below we briefly summarize methods based on their choice for $\mathbf{x}_2$, which leads to a regression or classification problem. 

\noindent\textbf{Regression.} 
In recent work Pathak \etal \cite{ContextEncoder} choose as surrogate label $\mathbf{x}_2$ a region of pixels in an image (\eg, the central patch) and use the remaining pixels in the image as $\mathbf{x}_1$. The model used for $p(\mathbf{x}_2|\mathbf{x}_1)$ is based on generative adversarial networks \cite{GAN14, DGAN16}. Other related work \cite{colorful,larsson2016learning} maps images to the Lab (luminance and opponent colors) space, and then uses the opponent colors as labels $\mathbf{x}_2$ and the luminance as $\mathbf{x}_1$.  Zhang \etal \cite{splitBrain} combine this choice to the opposite task of predicting the grayscale image from the opponent colors and outperform prior work. 


\noindent\textbf{Classification.} Doersch \etal and Noroozi \& Favaro \cite{context, noroozi2016} define a categorical problem where the surrogate labels are the relative positions of patches. Other recent works use as surrogate labels ego-motion \cite{agrawal15, tiedEgomotion}, temporal ordering in video \cite{shuffle, WatchingObjects}, sound \cite{ambientSound}, and physical interaction \cite{CuriousRobot}.


In contrast to these works, here we introduce a different formulation to arrive at a supervision signal. We define the \emph{counting} relationship \emph{``having the same number of visual primitives''} between two images. We use the fact that this relationship is satisfied by two identical images undergoing certain transformations, but not by two different images (although they might, with very low probability). Thus, we are able to assign a binary label (same or different number of visual primitives) to pairs of images. 
We are not aware of any other self-supervised method that uses this method to obtain the surrogate labels.
In contrast, Wang and Gupta \cite{wangVideo} impose relationships between triplets of different images obtained through tracking.
Notice that also Reed \etal \cite{reed15} exploit an explicit relationship between features. However, they rely on labeling that would reveal the relationship between different input images. Instead, we only exploit the structure of images and relate different parts of the same image to each other. 
Due to the above counting relationship our work relates also to \emph{object counting}, which we revise here below.

\noindent\textbf{Object counting.}
In comparison to other semantic tasks, counting has received little attention in the computer vision community. Most effort has been devoted to counting just one category 
and only recently it was applied to multiple categories in a scene. 
Counting is usually addressed as a supervised task, where a model is trained on annotated images. The counting prediction can be provided as an object density map \cite{Haroon15,Shao15,Victor10} or simply as the number of counted objects \cite{Antoni09,Everyday16}. There are methods to count humans in crowds \cite{Chan08, Crowd15_1, Crowd15_2}, cars \cite{cars16}, and penguins \cite{Penguins16}. Some recent works count common objects in the scene without relying on object localization \cite{Everyday16, EndToEnd17}.

In this work, we are not interested in the task of counting \emph{per se}. As mentioned earlier on, counting is used as a pretext task to learn a representation. Moreover, we do not use labels about the number of objects during training.


\section{Transforming Images to Transform Features}

One way to characterize a feature of interest is to describe how it should vary as a function of changes in the input data. For example, a feature that counts visual primitives should not be affected by scale, 2D translation, and 2D rotation changes of the input image. Other relationships might indicate instead that a feature should increase its values as a result of some transformation of the input. For example, the magnitude of the feature for counting visual primitives applied to half of an image should be smaller than when applied to the whole image. In general, we propose to learn a deep representation by using the known relationship between input and output transformations as a supervisory signal. To formalize these concepts, we first need to introduce some notation.

Let us denote a color image with $\mathbf{x}\in \real^{m \times n \times 3}$, where $m\times n$ is the size in pixels and there are $3$ color channels (RGB).
We define a family of image transformations ${\cal G} \triangleq \{G_1,\dots,G_J\}$, where $G_j:\real^{m\times n \times 3}\mapsto \real^{p\times q \times 3}$, with $j=1,\dots,J$, that take images $\mathbf{x}$ and map them to images of $p\times q$ pixels.
Let us also define a feature $\phi:\real^{p\times q \times 3} \mapsto \real^k$ mapping the transformed image to some $k$-dimensional vector. Finally, we define a feature transformation $g:\real^{k}\times\dots\times\real^k\mapsto \real^{k}$ that takes $J$ features and maps them to another feature.
Given the image transformation family $\cal G$ and $g$, we learn the feature $\phi$ by using the following relationship as an \emph{artificial supervisory signal} 
\begin{align}
g\left(\phi(G_1\circ \mathbf{x}),\dots,\phi(G_J\circ \mathbf{x})\right) = \mathbf{0}\quad\quad\quad \forall \mathbf{x}. 
\end{align}


In this work, the transformation family consists of the \emph{downsampling} operator $D$, with a downsampling factor of 2, and the \emph{tiling} operator $T_j$, where $j=1,\dots,4$, which extracts the $j-$th tile from a $2\times 2$ grid of tiles. Notice that these two transformations produce images of the same size. 
Thus, we can set ${\cal G} \equiv \{D,T_1,\dots,T_4\}$. We also define our desired relation between counting features on the transformed images as $g(\mathbf{d},\mathbf{t_1},\dots,\mathbf{t_4}) = \mathbf{d}-\sum_{j=1}^4 \mathbf{t_j}$. This can be written explicitly as
\begin{align}
\phi(D \circ \mathbf{x}) = \sum_{j=1}^4 \phi(T_j \circ \mathbf{x}).
\label{eq:constraint}
\end{align}
We use eq.~\eqref{eq:constraint} as our main building block to learn features $\phi$ that can count visual primitives.

This relationship has a bearing also on \emph{equivariance} \cite{LencV15}. Equivariance, however, is typically defined as the property of a given feature. In this work we invert this logic by fixing the transformations and by finding instead a representation satisfying those transformations.
Moreover, equivariance has restrictions on the type of transformations applied to the inputs and the features.

Notice that we have no simple way to control the scale at which our counting features work. It could count object parts, whole objects, object groups, or any combination thereof. This choice might depend on the number of elements of the counting vector $\phi$, 
on the loss function used for training, and on the type of data used for training. 

\section{Learning to Count}

\begin{figure}[t]
\begin{center}
\includegraphics[width=1\linewidth,trim={0cm .7cm .5cm 0cm},clip]{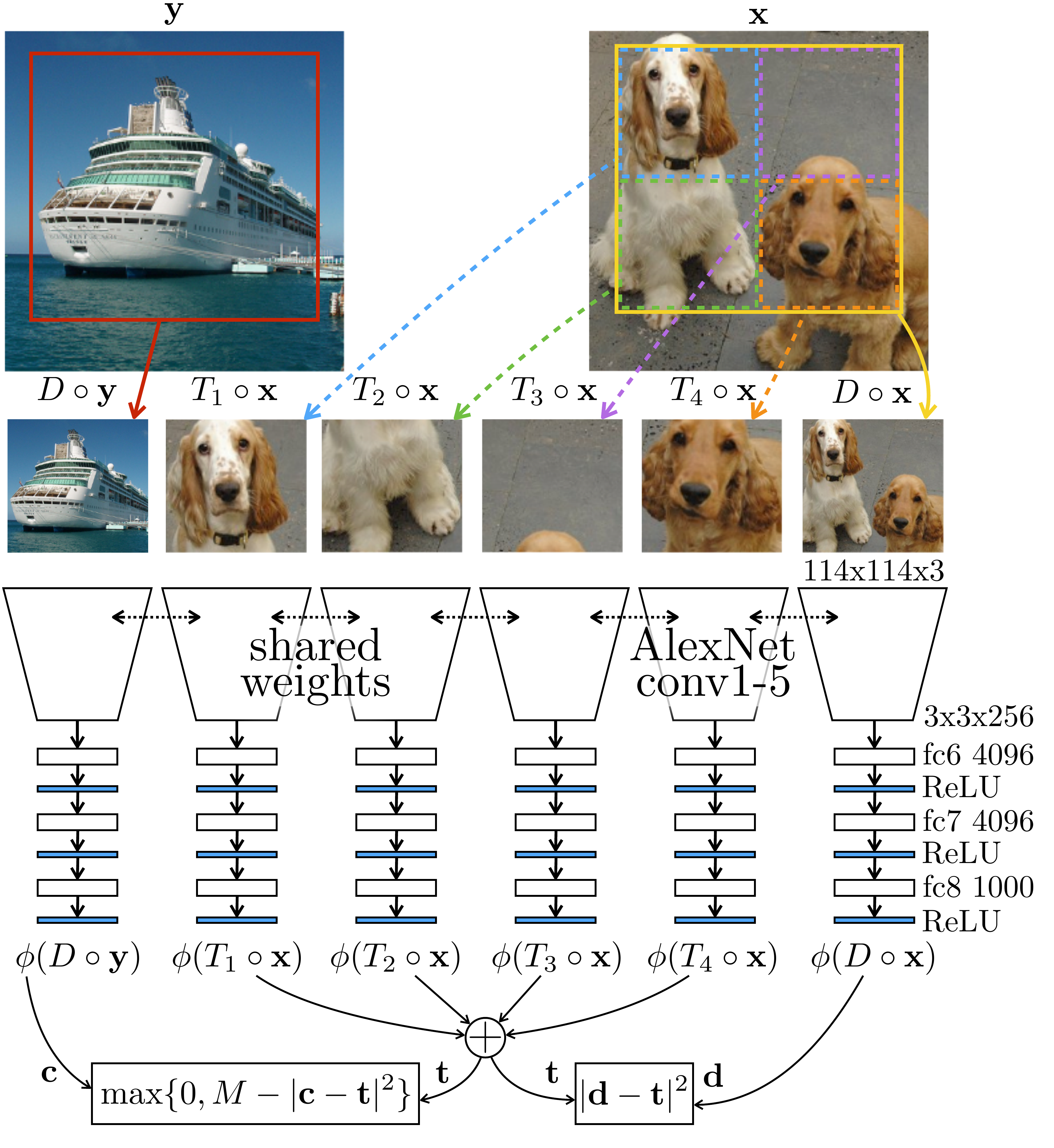}
\end{center}
   \caption{\textbf{Training AlexNet to learn to count.} The proposed architecture uses a siamese arrangement so that we simultaneously produce features for $4$ tiles and a downsampled image. We also compute the feature from a randomly chosen downsampled image ($D\circ \mathbf{y}$) as a contrastive term.}
\label{fig:siameseAlexNet}
\end{figure}

We use convolutional neural networks to obtain our representation.  In principle, our network could be trained with color images $\mathbf{x}$ from a large database (\eg, ImageNet \cite{ImageNet} or COCO \cite{COCO})  using an $l_2$ loss based on eq.~\eqref{eq:constraint}, for example,
\begin{align}
\textstyle\ell(\mathbf{x}) = \left| \phi(D\circ \mathbf{x}) - \sum_{j=1}^4 \phi(T_j\circ \mathbf{x})\right|^2.
\label{eq:l2loss}
\end{align}
\hamedcom{should we move the least effort bias paragraph to here before introducing the loss?}
However, this loss has $\phi(\mathbf{z}) = \mathbf{0}$, $\forall \mathbf{z}$, as its trivial solution. To avoid such a scenario, we use a contrastive loss~\cite{contrstive}, where we also enforce that the counting feature should be different between two randomly chosen different images. Therefore, for any $\mathbf{x}\neq \mathbf{y}$, we would like to minimize
\begin{align}
\label{eq:contrastive}
\ell_\text{con}(\mathbf{x},\mathbf{y}) \textstyle = \left| \phi(D\circ \mathbf{x}) - \sum_{j=1}^4 \phi(T_j\circ \mathbf{x})\right|^2\mspace{60mu}\\
\textstyle+ \max\left\{0, M - \left| \phi(D\circ \mathbf{y})- \sum_{j=1}^4 \phi(T_j\circ \mathbf{x})\right|^2 \right\}\notag
\end{align}
where in our experiments the constant scalar $M=10$.

\begin{figure}[t]
\begin{center}
\includegraphics[width=.9\linewidth,trim={0 0 0 0},clip]{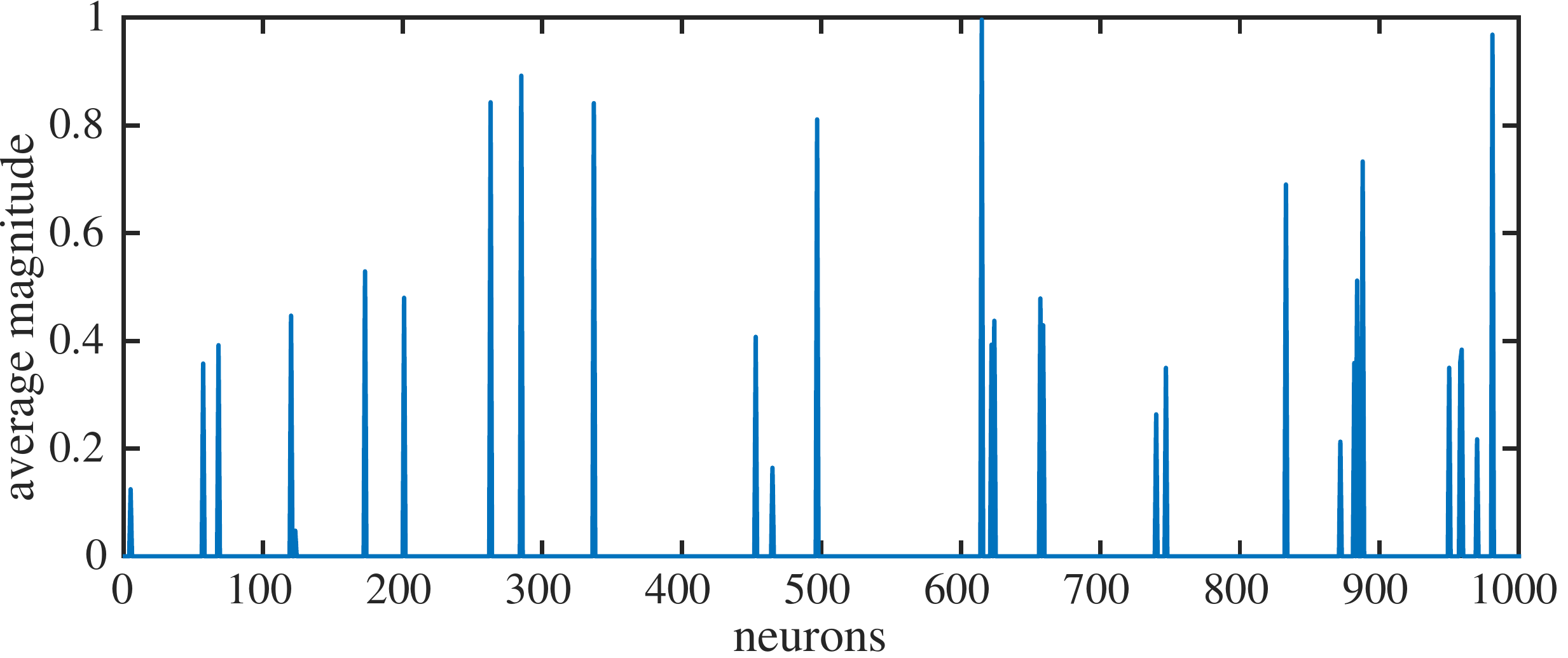}
\end{center}
   \caption{{\textbf{Average response of our trained network on the ImageNet validation set.}} Despite its sparsity ($30$ non zero entries), the hidden representation in the trained network performs well when transferred to the classification, detection and segmentation tasks.} 
\label{fig:activeneurons}
\vspace{-.1in}
\end{figure}

\noindent\textbf{Least effort bias.}
A bias of the system is that it can easily satisfy the constraint~\eqref{eq:l2loss} by learning to count as few visual primitives as possible. Thus, many entries of the feature mapping may collapse to zero.
This effect is observed in the final trained network. In Fig.~\ref{fig:activeneurons}, we show the average of features computed over the ImageNet validation set. There are only $30$  and $44$ non zero entries out of $1000$ after training on ImageNet and on COCO respectively. Despite the sparsity of the features, our transfer learning experiments show that the features in the hidden layers (\texttt{conv1-conv5}) perform very well on several benchmarks.
In our formulation~\eqref{eq:contrastive}, the contrastive term limits the effects of the least effort bias. Indeed, features that count very few visual primitives cannot differentiate much the content across different images. Therefore, the contrastive term will introduce a tradeoff that will push features towards counting as many primitives as is needed to differentiate images from each other.

\noindent {\bf Network architecture.} In principle, the choice of the architecture is arbitrary. For ease of comparison with state-of-the-art methods when transferring to classification and detection tasks,  we adopt the AlexNet architecture \cite{AlexNet12} as commonly done in other self-supervised learning methods. We use the first 5 convolutional layers from AlexNet followed by three fully connected layers ($(3\times3\times256)\times4096$, $4096\times4096$, and $4096\times1000$), and ReLU units. Note that $1000$ is the number of elements that we want to count. We use ReLU in the end since we want the counting vector to be all positive. Our input is $114\times114$ pixels to handle smaller tiles. Because all the features are the same, training with the loss function in eq.~\ref{eq:contrastive} is equivalent to training a 6-way siamese network, as shown in Fig.~\ref{fig:siameseAlexNet}.

\section{Experiments}


We first present the evaluations of our learned representation in the standard transfer learning benchmarks. Then, we perform ablation studies on our proposed method to show quantitatively  the impact of our techniques to prevent poor representations. Finally, we analyze the learned representation through some quantitative and qualitative experiments to get a better insight into what has been learned. 
We call the activation of the last layer of our network, on which the loss~\eqref{eq:contrastive} is defined, the \emph{counting vector}. We evaluate whether each unit in the counting vector is counting some visual primitive or not.
Our model is based on AlexNet \cite{AlexNet12} in all experiments.
In our tables we use boldface for the top performer and underline the second top performer.\\
\noindent\textbf{Implementation Details.~}
We use caffe \cite{jia2014caffe} with the default weight regularization settings to train our network. The learning rate is set to be quite low to avoid divergence. We begin with a learning rate of $10^{-4}$ and drop it by a factor of $0.9$ every $10K$ iterations. 
An important step is to normalize the input by subtracting the mean intensity value and dividing the zero-mean images by their standard deviation.

\subsection{Transfer Learning Evaluation} \label{ssec:tr}
We evaluate our learned representation on the \textit{detection}, \textit{classification}, and \textit{segmentation} tasks on the \textbf{PASCAL} dataset as well as the \textit{classification} task on the \textbf{ImageNet} dataset. We train our counting network on the 1.3M images from the training set of ImageNet. We use images of $114 \times 114$ pixels as input. Since we transfer only the convolutional layers, it has no effect on the transferred models and evaluation.
A new version of \cite{noroozi2016} has been released \cite{norooziArXiv2017}, where the standard AlexNet is used for transfer learning. All the numbers in our comparisons are from that version.

\begin{table}[t]
\footnotesize
\centering
\begin{adjustbox}{width=.48\textwidth}
\begin{tabular}{@{}l@{\hspace{2.3em}} c@{\hspace{2.1em}} c@{\hspace{2.1em}}c@{\hspace{2.1em}}  c@{}}
\toprule 
\textbf{Method}  & \textbf{Ref} & \textbf{Class.} & \textbf{Det.} & \textbf{Segm.}\\
\midrule 
Supervised~\cite{AlexNet12}   & \cite{colorful} & 				79.9 & 				56.8 & 48.0 \\
Random  	&  \cite{ContextEncoder} & 							53.3 & 				43.4 & 19.8 \\
\hline
Context~\cite{context}  	& \cite{krahenbuhl2015data}  & 		55.3 & 				46.6 & - \\
Context~\cite{context}$^*$  	& \cite{krahenbuhl2015data}  & 	65.3 & 				51.1 & - \\
Jigsaw~\cite{norooziArXiv2017}    & \cite{norooziArXiv2017} &  \underline{67.6}& 		\textbf{53.2}& \underline{37.6} \\
ego-motion~\cite{agrawal15}  	& \cite{agrawal15}  & 			52.9 & 				41.8 & - \\
ego-motion~\cite{agrawal15}$^*$  	& \cite{agrawal15}  & 		54.2 & 				43.9 & - \\
Adversarial~\cite{advLearning}$^*$ & \cite{advLearning}  & 		58.6 & 				46.2 & 34.9 \\
ContextEncoder~\cite{ContextEncoder} & \cite{ContextEncoder}  & 56.5 & 				44.5 & 29.7 \\
Sound~\cite{ambientSound}   &  \cite{splitBrain} & 				54.4 & 				44.0 & - \\
Sound~\cite{ambientSound}$^*$   &  \cite{splitBrain} & 			61.3 & 				- & - \\
Video~\cite{wangVideo}   &  \cite{krahenbuhl2015data}  &		62.8 & 				47.4 &  - \\
Video~\cite{wangVideo}$^*$  & \cite{krahenbuhl2015data}  & 		63.1 & 				47.2 & - \\
Colorization~\cite{colorful}$^*$  &  \cite{colorful} & 			65.9 & 				46.9 & 35.6 \\
Split-Brain~\cite{splitBrain}$^*$     & \cite{splitBrain} & 	67.1 & 				46.7 & 36.0 \\
ColorProxy~\cite{larsson2017colorproxy} & ~\cite{larsson2017colorproxy} & 65.9 & - & \textbf{38.0} \\
WatchingObjectsMove~\cite{WatchingObjects} &~\cite{WatchingObjects}   & 61.0 & \underline{52.2} & - \\
Counting  &   & 										\textbf{67.7}&  	 51.4 & 36.6 \\
\bottomrule
\end{tabular}
\end{adjustbox}
 \caption{{\textbf{Evaluation of transfer learning on PASCAL.}} Classification and detection are evaluated on PASCAL VOC 2007 in the frameworks introduced in \cite{krahenbuhl2015data} and \cite{girshickICCV15fastrcnn} respectively. Both tasks are evaluated using mean average precision (mAP) as a performance measure. Segmentation is evaluated on PASCAL VOC 2012 in the framework of \cite{fcn}, which reports mean intersection over union \hamed{(mIoU)}. (*) denotes the use of the data initialization method \cite{krahenbuhl2015data}.
}\label{tbl:voc}
\end{table}

\subsubsection{Fine-tuning on PASCAL} \label{sssec:VOC2007}
In this set of experiments, we fine-tune our network on the PASCAL VOC 2007 and VOC 2012 datasets, which are standard benchmarks for representation learning. Fine-tuning is based on established frameworks for object classification~\cite{krahenbuhl2015data}, detection~\cite{girshickICCV15fastrcnn} and segmentation~\cite{fcn} tasks.\hamedcom{should we name them?} The classification task is a multi-class classification problem, which predicts the presence or absence of $20$ object classes. The detection task involves locating objects by specifying a bounding box around them. Segmentation assigns the label of an object class to each pixel in the image.
As shown in Table~\ref{tbl:voc}, we either outperform previous methods or achieve the second best performance.
Notice that while classification and detection are evaluated on VOC 2007, segmentation is evaluated on VOC 2012. 
Unfortunately, we did not get any performance boost when using the  method of Kr\"{a}henb\"{u}hl \etal~\cite{krahenbuhl2015data}. 

\begin{table}[t]
\begin{adjustbox}{width=.48\textwidth}
\begin{tabular}{@{}l  c  c  c   c c c@{}}
\toprule
\textbf{Method}   & \textbf{conv1} & \textbf{conv2} & \textbf{conv3} & \textbf{conv4} & \textbf{conv5}\\
\midrule
Supervised~\cite{AlexNet12}  & 			19.3 & 36.3 & 44.2 & 48.3 & 50.5\\
Random & 11.6 & 17.1 & 16.9 & 16.3 & 14.1 \\
\hline
Context~\cite{context}     & 			16.2 & 23.3 & 30.2 & 31.7 & 29.6\\ 
Jigsaw~\cite{norooziArXiv2017} &   	\textbf{18.2}& 28.8 & 34.0 & \underline{33.9} & 27.1 \\
ContextEncoder~\cite{ContextEncoder}  & 	14.1 & 20.7 & 21.0 & 19.8 & 15.5\\
Adversarial~\cite{advLearning}  & 		17.7 & 24.5 & 31.0 & 29.9 & 28.0  \\
Colorization~\cite{colorful}    & 		12.5 & 24.5 & 30.4 & 31.5 & \underline{30.3} \\
Split-Brain~\cite{splitBrain}      & 	17.7 & \underline{29.3} & \textbf{35.4} & \textbf{35.2} & \textbf{32.8} \\
Counting & \underline{18.0} & \textbf{30.6} & \underline{34.3} & 32.5 & 25.7\\
\bottomrule
\end{tabular}
\end{adjustbox}
 \caption{{\textbf{ImageNet classification with a linear classifier.}} We use the publicly available code and configuration of \cite{colorful}. Every column shows the top-1 accuracy of AlexNet on the classification task. The learned weights from \texttt{conv1} up to the displayed layer are frozen. The features of each layer are spatially resized until there are fewer than 9K dimensions left. A fully connected layer followed by softmax is trained on a 1000-way object classification task.  
 } \label{tbl:imagenet_lin}
\end{table}

\begin{table}[t]
\footnotesize
\centering
\begin{tabular}{@{}l@{\hspace{2.8em}}  c@{\hspace{1.5em}}  c@{\hspace{1.5em}}   c@{\hspace{1.5em}} c@{\hspace{1.5em}} c@{}}
\toprule
\textbf{Method}   & \textbf{conv1} & \textbf{conv2} & \textbf{conv3} & \textbf{conv4} & \textbf{conv5}\\
\midrule
Places labels~\cite{places}  & 			22.1 & 35.1 & 40.2 & 43.3 & 44.6\\
ImageNet labels~\cite{AlexNet12}  & 	22.7 & 34.8 & 38.4 & 39.4 & 38.7\\
Random    &  							15.7 & 20.3 & 19.8 & 19.1 & 17.5 \\
\hline
Context~\cite{context}  & 				19.7 & 26.7 & 31.9 & 32.7 & \underline{30.9}\\ 
Jigsaw~\cite{norooziArXiv2017} &  \underline{23.0} & \underline{31.9} & \underline{35.0} & \underline{34.2} & 29.3 \\
Context encoder~\cite{ContextEncoder} & 18.2 & 23.2 & 23.4 & 21.9 & 18.4 \\
Sound~\cite{ambientSound}& 				19.9 & 29.3 & 32.1 & 28.8 & 29.8 \\
Adversarial~\cite{advLearning} & 		22.0 & 28.7 & 31.8 & 31.3 & 29.7  \\
Colorization~\cite{colorful}    & 		16.0 & 25.7 & 29.6 & 30.3 & 29.7 \\
Split-Brain~\cite{splitBrain}    & 		21.3 & 30.7 & 34.0 & 34.1 & \textbf{32.5} \\
Counting & 						\textbf{23.3} & \textbf{33.9} & \textbf{36.3} & \textbf{34.7} & 29.6 \\
\bottomrule
\end{tabular}
 \caption{{\textbf{Places classification with a linear classifier.}} \hamed{We use the same setting as in Table~\ref{tbl:imagenet_lin} except that to evaluate generalization across datasets, the model is pretrained on ImageNet (with no labels) and then tested with frozen layers on Places (with labels). The last layer has 205 neurons for scene categories.}
} \label{tbl:places_lin}
\end{table}

\subsubsection{Linear Classification on Places and ImageNet} \label{sssec:ImageNetCls}
As introduced by Zhang \etal~\cite{colorful}, we train a linear classifier on top of the frozen layers on ImageNet~\cite{ImageNet} and Places~\cite{places} datasets. 
The results of these experiments are shown in Tables~\ref{tbl:imagenet_lin} and \ref{tbl:places_lin}. 
Our method achieves a performance comparable to the other state-of-the-art methods on the ImageNet dataset and shows a significant improvement on the Places dataset. Training and testing a method on the same dataset type, although with separate sets and no labels, may be affected by \emph{dataset bias}. To have a better assessment of the generalization properties of all the competing methods, \hamed{we suggest (as shown in Table~\ref{tbl:places_lin}) using the ImageNet dataset for training and the Places benchmark for testing (or vice versa).}
Our method archives state-of-the-art results with the \texttt{conv1-conv4} layers on the Places dataset.
Interestingly, the performance of our \texttt{conv1} layer is even \textbf{higher than the one obtained with supervised learning} when trained either on ImageNet or Places labels. The values for all the other methods in Tables~\ref{tbl:imagenet_lin} and \ref{tbl:places_lin} are taken form~\cite{splitBrain} except for ~\cite{norooziArXiv2017}, which we report for the first time.


\subsection{Ablation Studies}\label{sec:trivial}

\begin{table}[t]
\centering
\begin{adjustbox}{width=.48\textwidth}
\begin{tabular}{@{} l  c  c  c  c @{}}
\toprule
\textbf{Interpolation}   & \textbf{Training} & \textbf{Color} & \textbf{Counting } & \textbf{Detection} \\
 $\quad$\textbf{method}   & \textbf{size} & \textbf{space} & \textbf{dimension} & \textbf{performance} \\
\midrule
 $\quad$Mixed & 1.3M  & RGB/Gray & 20 & 50.9\\
\midrule
 $\quad$Mixed & 128K  & Gray & 1000 & 44.9\\
 $\quad$Mixed & 512K  & Gray & 1000 & 49.1\\
\midrule
 $\quad$Mixed & 1.3M  & RGB & 1000 & 48.2\\
 $\quad$Mixed & 1.3M  & Gray & 1000 & 50.4\\
\midrule
 $\quad$Linear & 1.3M & RGB/Gray & 1000 & 48.4\\
 $\quad$Cubic & 1.3M  & RGB/Gray & 1000 & 48.9 \\
 $\quad$Area & 1.3M  & RGB/Gray & 1000& 49.2\\ 
 $\quad$Lanczos & 1.3M  & RGB/Gray & 1000 & 47.3 \\
\midrule
 $\quad$Mixed & 1.3M  & RGB/Gray & 1000 & \textbf{51.4}\\
\bottomrule
\end{tabular}
\end{adjustbox}
 \caption{\textbf{Ablation studies.} We train the counting task  under different interpolation methods, training size/color, and feature dimensions, and compare the performance of the learned representations on the detection task on the PASCAL VOC 2007 dataset. 
 } \label{tbl:ablation_det}
\end{table}

To show the effectiveness of our proposed method, in Table~\ref{tbl:ablation_det} we compare its performance on the detection task on PASCAL VOC 2007 under different training scenarios. 
The first three rows illustrate some simple comparisons based on feature and dataset size.
The first row shows the impact of the counting vector length. As discussed earlier on, the network tends to generate sparse counting features. We train the network on ImageNet with only $20$ elements in the counting vector. This leads to a small drop in the performance, thus showing little sensitivity with respect to feature length. We also train the network with a smaller set of training images. The results show that our method is sensitive to the size of the training set. This shows that the counting task is non-trivial and requires a large training dataset.

The remaining rows in Table~\ref{tbl:ablation_det} illustrate a more advanced analysis of the counting task.
An important part of the design of the learning procedure is the identification of trivial solutions, \ie, solutions that would not result in a useful representation and that the neural network could converge to. By identifying such \emph{trivial learning} scenarios, we can provide suitable countermeasures. We now discuss possible \emph{shortcuts} that the network could use to solve the counting task and also the techniques that we use to avoid them.

A first potential problem is that the neural network learns trivial features such as low-level texture statistics histograms. For example, a special case is color histograms. This representation is undesirable because it would be semantically agnostic (or very weak) and therefore we would not expect it to transfer well to classification and detection. In general, these histograms would not satisfy eq.~\eqref{eq:constraint}. However, if the neural network could tell tiles apart from downsampled images, then it could apply a customized scaling factor to the histograms in the two cases and satisfy eq.~\eqref{eq:constraint}. In other words, the network might learn the following degenerate feature
\begin{align}
\phi(\mathbf{z}) = \begin{cases}\frac{1}{4}\text{hist}(\mathbf{z})&\text{if }\mathbf{z}\text{ is a tile}\\
~~\text{hist}(\mathbf{z})&\text{if }\mathbf{z}\text{ is downsampled}.
\end{cases}
\end{align}
Notice that this feature would satisfy the first term in eq.~\eqref{eq:constraint}. The second (contrastive) term would also be easily satisfied since different images have typically different low-level texture histograms.
We discuss below scenarios when this might happen and present our solutions towards reducing the likelihood of trivial learning.

\begin{table}[t]
\begin{adjustbox}{width=.48\textwidth}
\begin{tabular}{  @{\hspace{.5em}}l@{\hspace{.5em}}   | c  c  c  c c |c }
\noalign{\hrule height 1pt}
train$\backslash$test& Linear  & Cubic & Area & Lanczos & Mixed & std\\
\noalign{\hrule height .7pt}
Linear & \textbf{0.33}  &  0.63 &  0.33 &   0.65 &  0.48  & 0.18\\
Cubic & 0.79 &   \textbf{0.25}  &   0.78 &   0.22 &   0.52 & 0.32\\
Area & 0.32  &  0.85  &  \textbf{0.31}  &  0.95 &   0.50 & 0.34\\
Lanczos & 1.023  &  0.31   & 1.02  &  \textbf{0.19} &  0.58 & 0.45  \\ 
Mixed & 0.36  & \textbf{0.29}  &  0.37  &  0.30  &  0.34 & \textbf{0.04}\\\noalign{\hrule height 1pt}
\end{tabular}
\end{adjustbox}
 \caption{{\textbf{Learning the downsampling style.}} The first column and row show the downsampling methods used during the training and testing time respectively. The values in the first block show the pairwise error metric in eq.~\eqref{eq:errormetric} between corresponding  downsampling methods.  The last column shows the standard deviation of the error metric across different downsampling methods at test time. 
 } \label{tbl:down_samp}
\end{table}

\noindent\textbf{The network recognizes the downsampling style.} During training, we randomly crop a $224\times 224$ region from a $256\times 256$ image. Next, we downsample the whole image by a factor of $2$. The downsampling style, \eg, bilinear, bicubic, and Lanczos, may leave artifacts in images that the network may learn to recognize. To make the identification of the downsampling method difficult, at each \emph{stochastic gradient descent} iteration, we randomly pick either the bicubic,  bilinear, lanczos, or the area method as defined in OpenCV \cite{opencv}. As shown in Table~\ref{tbl:ablation_det}, the randomization of different downsampling methods significantly improves the detection performance by at least $2.2\%$. 

In Table~\ref{tbl:down_samp}, we perform another experiment that clearly shows that network learns the downsampling style. We train our network by using only one downsampling method. Then, we test the network on the pretext task by using only one (possibly different) method. If the network has learned to detect the downsampling method, then it will perform poorly at test time when using a different one. As an error metric, we use the first term in the loss function normalized by the average of the norm of the feature vector. More precisely, the error when the network is trained with the $i$-th downsampling style and tested on the $j$-th one is
\begin{align}
\mathbf{e}_{ij} = \frac{\sum_{\mathbf{x}} \left|  \sum_{p=1}^4 \phi^i \left( T_p \circ \mathbf{x}) - \phi^i(D^j \circ \mathbf{x}\right) \right|^2}{\sum_{\mathbf{x}} \left| \phi^{i} \left( D^i\circ \mathbf{x}\right)\right|^2}
\label{eq:errormetric}
\end{align}
where $\phi^{i}$ denotes the counting vector of the network trained with the $i$-th downsampling method. $D^{j}$ denotes the downsampling transformation using the $j$-th  method. $T_p$ is the tiling transformation that gives the $p$-th tile. 

Table~\ref{tbl:down_samp} collects all the computed errors. \hamed{The element in row $i$ and column $j$}
shows the pairwise error metric $\mathbf{e}_{ij}$. The last column shows the standard deviation of  this error metric across different downsampling methods. A higher value means that the network is sensitive to the downsampling method. This experiment clearly shows that the network learns the downsampling style. Another observation that can be made based on the similarity of the errors, is that the pairs (linear, area) and (cubic, lanczos) \hamed{leave similar artifacts in downsampling.}

\noindent\textbf{The network recognizes chromatic aberration.} The presence of chromatic aberration and its undesirable effects on learning have been pointed out by Doersch \etal \cite{context}. Chromatic aberration is a relative shift between the color channels that increases in the outward radial direction. Hence, our network can use this property to tell tiles apart from the dowsampled images. In fact, tiles will have a strongly diagonal chromatic aberration, while the downsampled image will have a radial aberration. We already reduce its effect by choosing the central region in the very first cropping preprocessing. To further reduce its effect, we train the network with both color and grayscale images (obtained by \hamed{replicating the average color across all $3$ channels).}
In training, we randomly choose color images $33\%$ of the time and grayscale images $67\%$ of the time. This choice is consistent across all the terms in the loss function (\ie, all tiles and downsampled images are either colored or grayscale). 
While this choice does not completely solve the issue, it does improve the performance of the model. We find that completely eliminating the color from images leads to a loss in performance in transfer learning (see Table~\ref{tbl:ablation_det}).

\begin{figure*}[th]
  \centering
  \begin{minipage}[b]{.245\textwidth}
        \includegraphics[width=1 \textwidth]{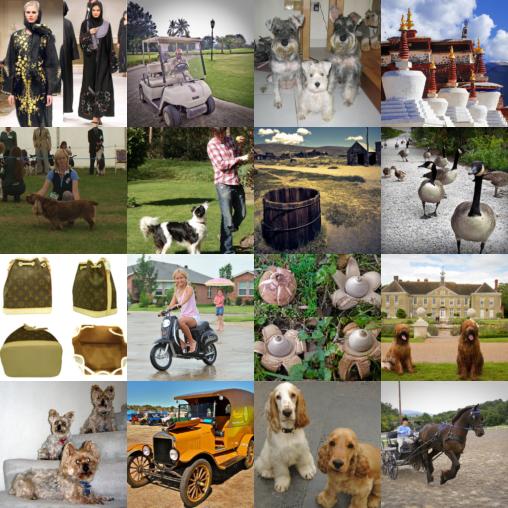}
        \vspace{-.5cm}
        \subcaption{}
   \end{minipage}
   \begin{minipage}[b]{.245\textwidth}
        \includegraphics[width=1 \textwidth]{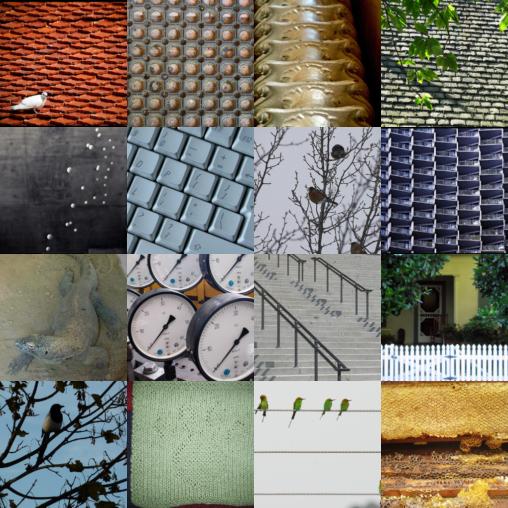}
        \vspace{-.5cm}
        \subcaption{}
   \end{minipage}    
   \begin{minipage}[b]{.245\textwidth}
        \includegraphics[width=1 \textwidth]{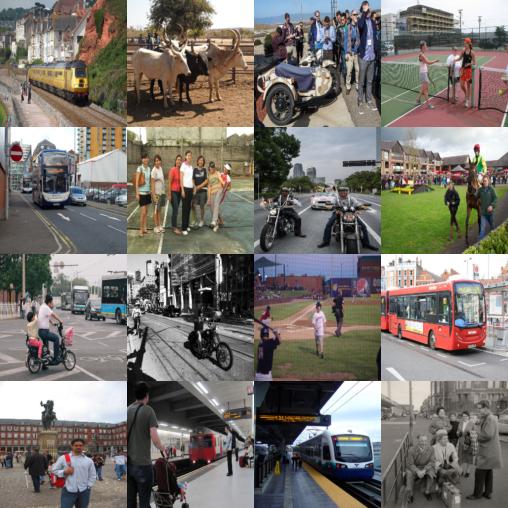}
        \vspace{-.5cm}
        \subcaption{}
   \end{minipage}
    \begin{minipage}[b]{.245\textwidth}
        \includegraphics[width=1 \textwidth]{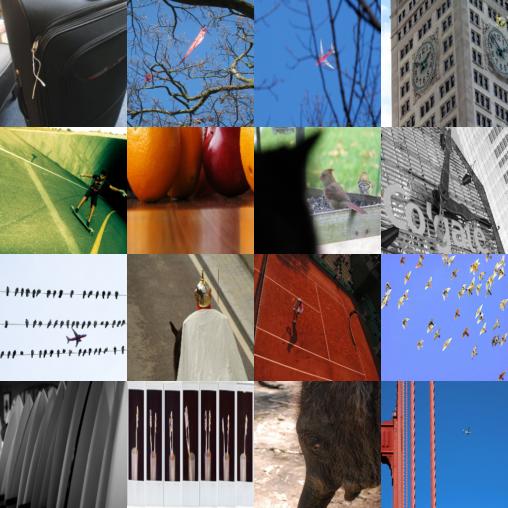}
        \vspace{-.5cm}
        \subcaption{}
   \end{minipage}    
\vspace{-.1in}
  \caption{{\textbf{Examples of activating/ignored images.}} (a) and (b) show the top 16 images with the highest and lowest counting feature magnitude from the validation set of ImageNet. (c) and (d) show the top 16 images with the highest and lowest counting feature magnitude from the test set of COCO. }
  \label{fig:high_low_norm}
\vspace{-.15in}
\end{figure*}

\subsection{Analysis} \label{ssec:analyses}

We use visualization and nearest neighbor search to see what visual primitives our trained network counts. Ideally, these visual primitives should capture high-level concepts like objects or object parts rather than low-level concepts like edges and corners. In fact, detecting simple corners will not go a long way in semantic scene understanding. To avoid dataset bias, \hamed{we train our model on ImageNet (with no labeles) and show the results on COCO dataset.}


\begin{figure}[t]
  \centering
      \includegraphics[width=0.212\linewidth]{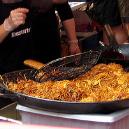}\hspace{.3cm}        
      \includegraphics[width=0.212\linewidth]{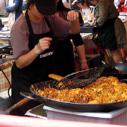}\hspace{.3cm}
      \includegraphics[width=0.212\linewidth]{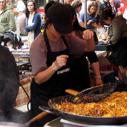}\hspace{.3cm}
      \includegraphics[width=0.212\linewidth]{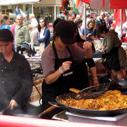}
  \caption{{\textbf{Image croppings of increasing size.}} The number of visual primitives should increase going from left to right.}
  \label{fig:crops}
\end{figure}

\begin{figure}[t]
  \centering
      \includegraphics[width=.99\linewidth]{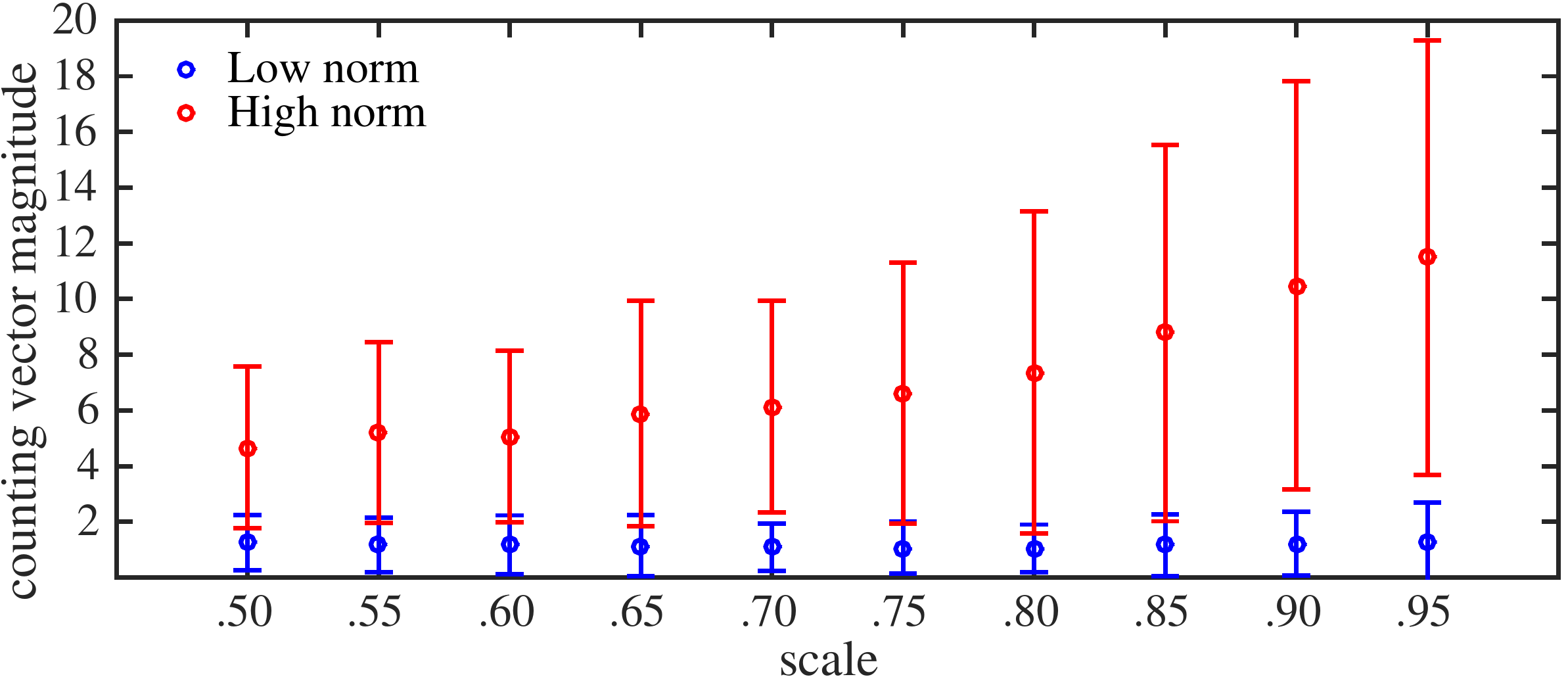}
  \caption{{\textbf{Counting evaluation on ImageNet.}} On the abscissa we report the scale of the cropped region and on the ordinate the corresponding average and standard deviation of the counting vector  magnitude.}
  \label{fig:low_hig_plot}
\vspace{-.1in}
\end{figure}

\subsubsection{Quantitative Analysis}

We illustrate quantitatively the relation between the magnitude of the counting vector and the number of objects. Rather than counting exactly the number of specific objects, we introduce a simple method to rank images based on how many objects they contain. The method is based on cropping an image with larger and larger regions which are then rescaled to the same size through downsampling (see Fig.~\ref{fig:crops}). We build two sets of $100$ images each. \hamed{We assign images yielding the highest and lowest feature magnitude into two different sets.}
We randomly crop 10 regions with an area between $50\%-95\%$ of each image and compute the corresponding counting vector. The mean and the standard deviation of the  counting vector magnitude of the cropped images for each set is shown in Fig~\ref{fig:low_hig_plot}. We observe that our feature does not count low-level texture, and is instead more sensitive to composite images. A better understanding of this observation needs futher investigation.


\fboxsep=0pt
\fboxrule=1pt

\begin{figure*}
  \centering
  \fcolorbox{red}{yellow}{\begin{minipage}[b]{.049\textwidth}
  		\includegraphics[width=1 \textwidth,trim={0 0 40.34cm 0},clip]{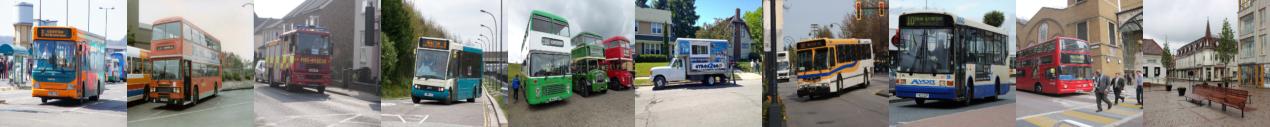}\vspace{-.4mm}\\
  		\includegraphics[width=1 \textwidth,trim={0 0 40.34cm 0},clip]{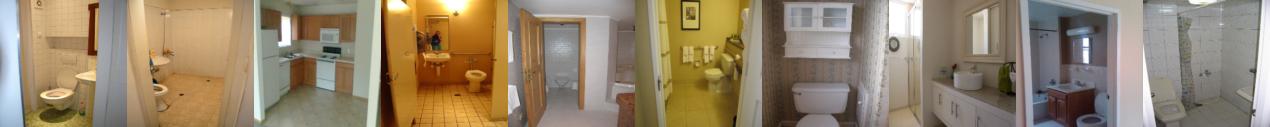}\vspace{-.4mm}\\
        \includegraphics[width=1 \textwidth,trim={0 0 40.34cm 0},clip]{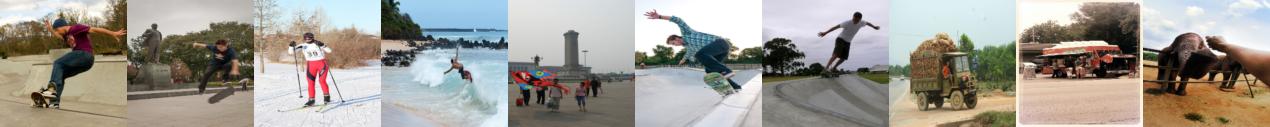}\vspace{-.4mm}\\
        \includegraphics[width=1 \textwidth,trim={0 0 40.34cm 0},clip]{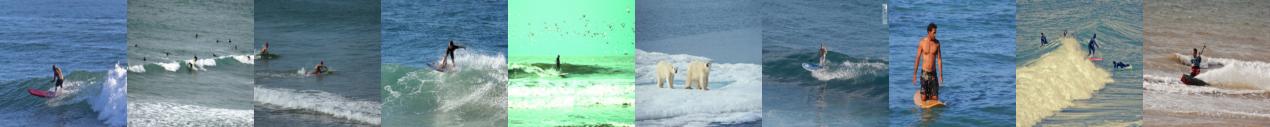}\vspace{-.4mm}\\
        \includegraphics[width=1 \textwidth,trim={0 0 40.34cm 0},clip]{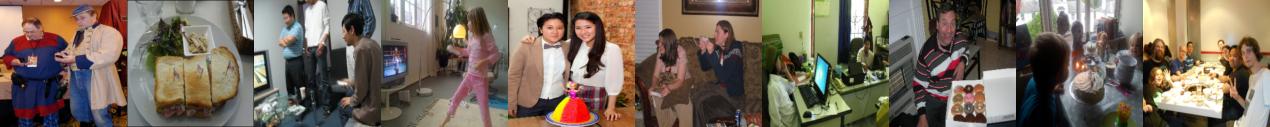}\vspace{-.4mm}\\
        \includegraphics[width=1 \textwidth,trim={0 0 40.34cm 0},clip]{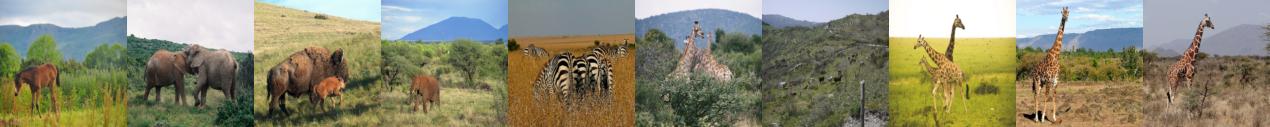}\vspace{-.4mm}\\
        \includegraphics[width=1 \textwidth,trim={0 0 40.34cm 0},clip]{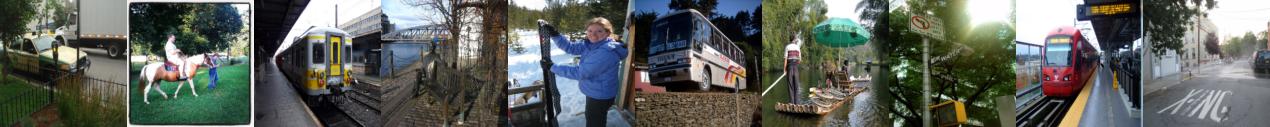}\vspace{-.4mm}\\
        \includegraphics[width=1 \textwidth,trim={0 0 40.34cm 0},clip]{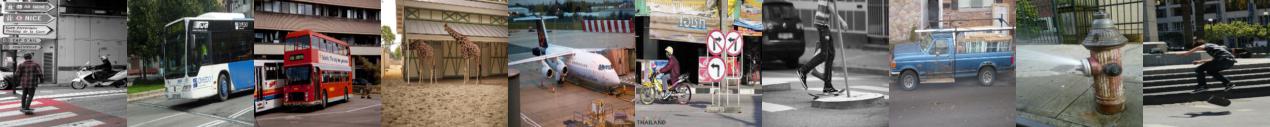}
   \end{minipage}}%
   \begin{minipage}[b]{.4427\textwidth}
   		\includegraphics[width=1 \textwidth,trim={4.48cm 0 0 0},clip]{mscoco_N_10_7}\vspace{-.4mm}\\
        \includegraphics[width=1 \textwidth,trim={4.48cm 0 0 0},clip]{mscoco_N_10_4}\vspace{-.4mm}\\
        \includegraphics[width=1 \textwidth,trim={4.48cm 0 0 0},clip]{mscoco_N_5_10}\vspace{-.4mm}\\
        \includegraphics[width=1 \textwidth,trim={4.48cm 0 0 0},clip]{mscoco_N_8_1}\vspace{-.4mm}\\
        \includegraphics[width=1 \textwidth,trim={4.48cm 0 0 0},clip]{mscoco_N_8_4}\vspace{-.4mm}\\
        \includegraphics[width=1 \textwidth,trim={4.48cm 0 0 0},clip]{mscoco_N_7_6}\vspace{-.4mm}\\
        \includegraphics[width=1 \textwidth,trim={4.48cm 0 0 0},clip]{mscoco_N_1_1}\vspace{-.4mm}\\
        \includegraphics[width=1 \textwidth,trim={4.48cm 0 0 0},clip]{mscoco_N_1_2}
   \end{minipage}
    \fcolorbox{red}{yellow}{\begin{minipage}[b]{.049\textwidth}
        \includegraphics[width=1 \textwidth,trim={0 0 40.34cm 0},clip]{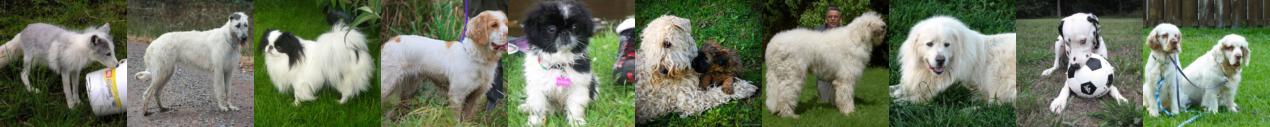}\vspace{-.4mm}\\
        \includegraphics[width=1 \textwidth,trim={0 0 40.34cm 0},clip]{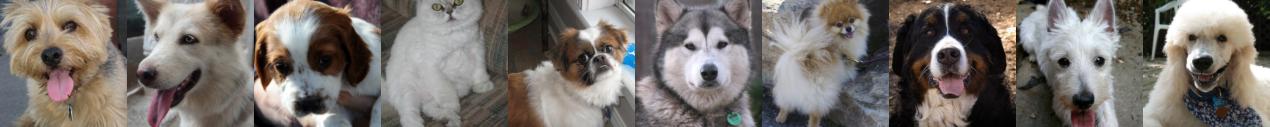}\vspace{-.4mm}\\
        \includegraphics[width=1 \textwidth,trim={0 0 40.34cm 0},clip]{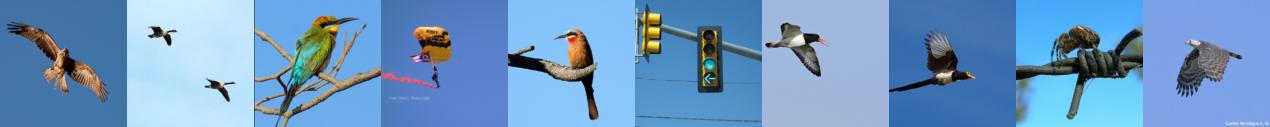}\vspace{-.4mm}\\
        \includegraphics[width=1 \textwidth,trim={0 0 40.34cm 0},clip]{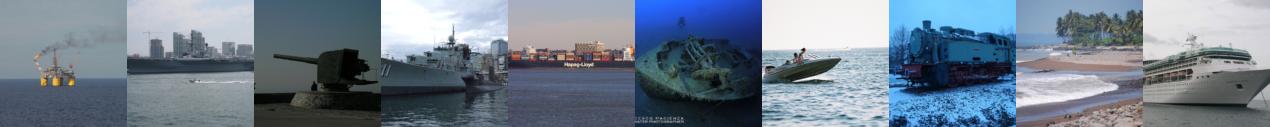}\vspace{-.4mm}\\
        \includegraphics[width=1 \textwidth,trim={0 0 40.34cm 0},clip]{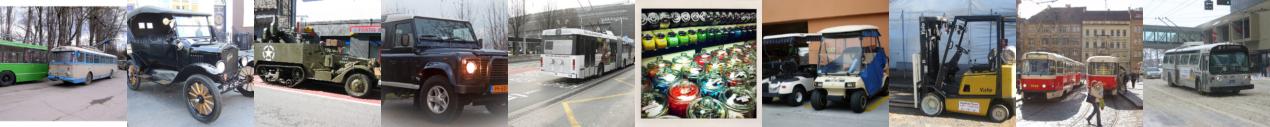}\vspace{-.4mm}\\
        \includegraphics[width=1 \textwidth,trim={0 0 40.34cm 0},clip]{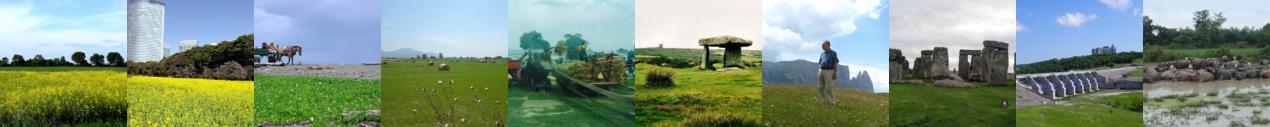}\vspace{-.4mm}\\
        \includegraphics[width=1 \textwidth,trim={0 0 40.34cm 0},clip]{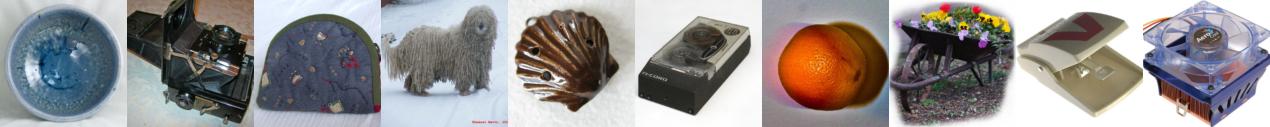}\vspace{-.4mm}\\
        \includegraphics[width=1 \textwidth,trim={0 0 40.34cm 0},clip]{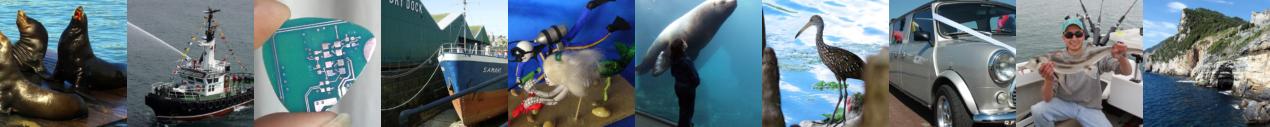}
   \end{minipage}}%
  \begin{minipage}[b]{.4427\textwidth}
        \includegraphics[width=1 \textwidth,trim={4.48cm 0 0 0},clip]{imagenet_N_2_8}\vspace{-.4mm}\\
        \includegraphics[width=1 \textwidth,trim={4.48cm 0 0 0},clip]{imagenet_N_10_5}\vspace{-.4mm}\\
        \includegraphics[width=1 \textwidth,trim={4.48cm 0 0 0},clip]{imagenet_N_1_6}\vspace{-.4mm}\\
        \includegraphics[width=1 \textwidth,trim={4.48cm 0 0 0},clip]{imagenet_N_7_4}\vspace{-.4mm}\\
        \includegraphics[width=1 \textwidth,trim={4.48cm 0 0 0},clip]{imagenet_N_10_10}\vspace{-.4mm}\\
        \includegraphics[width=1 \textwidth,trim={4.48cm 0 0 0},clip]{imagenet_N_10_7}\vspace{-.4mm}\\
        \includegraphics[width=1 \textwidth,trim={4.48cm 0 0 0},clip]{imagenet_N_1_3}\vspace{-.4mm}\\
        \includegraphics[width=1 \textwidth,trim={4.48cm 0 0 0},clip]{imagenet_N_3_2}
   \end{minipage}     
  \caption{{\textbf{Nearest neighbor retrievals.}} Left: COCO retrievals. Right: ImageNet retrievals. In both datasets, the leftmost column (with a red border) shows the queries and the other columns show the top matching images sorted with increasing Euclidean distance \hamed{in our counting feature space} from left to right. 
  On the bottom $3$ rows, we show the failure retrieval cases. Note that the matches share a similar content and scene outline. \hamedcom{can we draw a vertical line to separate query from results?}}
  \label{fig:NN}
\end{figure*}

\begin{figure*}
  \centering
  \begin{minipage}[b]{.245\textwidth}
        \includegraphics[width=1 \textwidth,trim={0cm 9cm 0cm 0cm},clip]{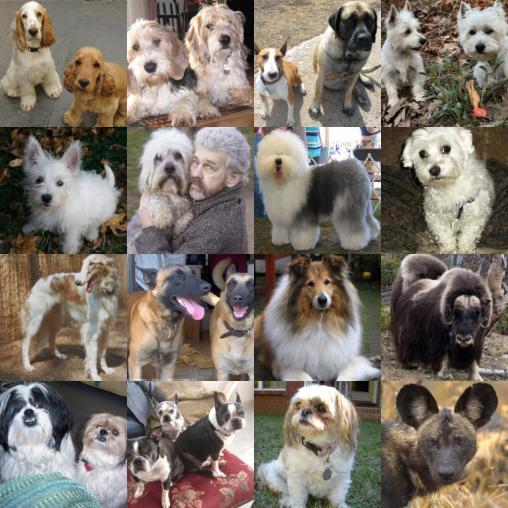}
   \end{minipage}
   \begin{minipage}[b]{.245\textwidth}
        \includegraphics[width=1 \textwidth,trim={0cm 9cm 0cm 0cm},clip]{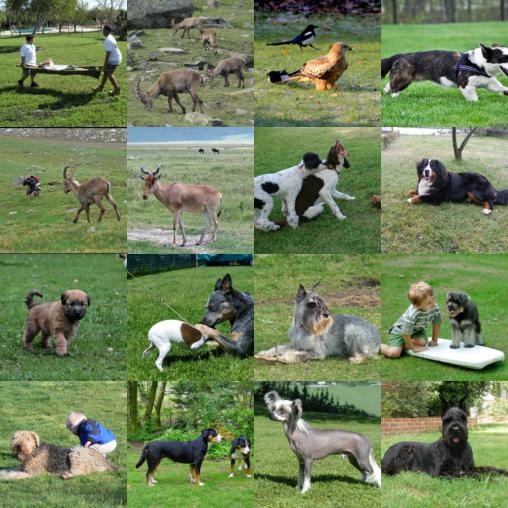}
   \end{minipage}    
   \begin{minipage}[b]{.245\textwidth}
        \includegraphics[width=1 \textwidth,trim={0cm 9cm 0cm 0cm},clip]{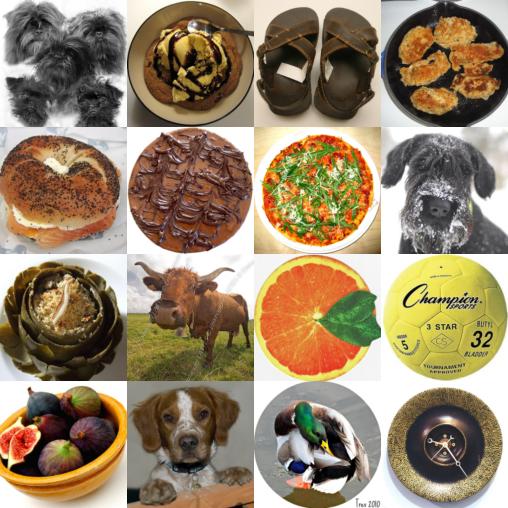}
   \end{minipage}
    \begin{minipage}[b]{.245\textwidth}
        \includegraphics[width=1 \textwidth,trim={0cm 9cm 0cm 0cm},clip]{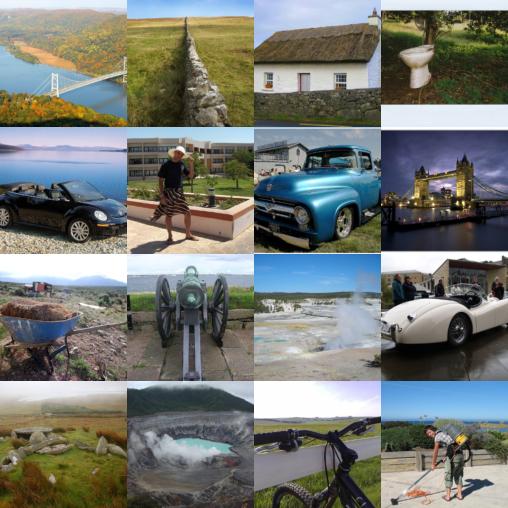}
   \end{minipage}\vspace{.5mm}\\
   \begin{minipage}[b]{.245\textwidth}
        \includegraphics[width=1 \textwidth,trim={0cm 9cm 0cm 0cm},clip]{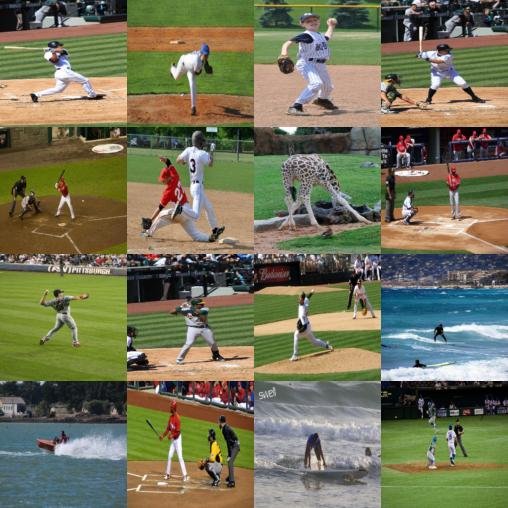}
   \end{minipage}
   \begin{minipage}[b]{.245\textwidth}
        \includegraphics[width=1 \textwidth,trim={0cm 9cm 0cm 0cm},clip]{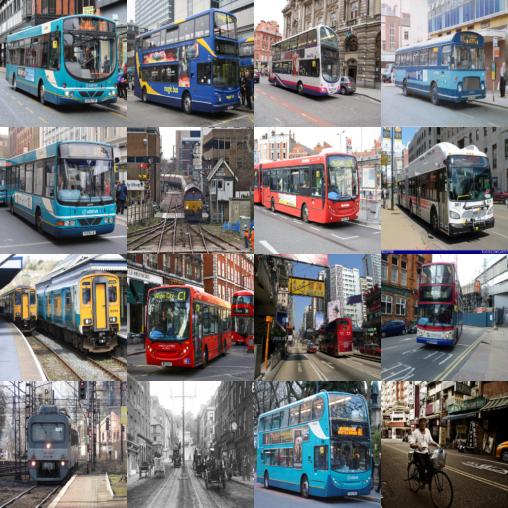}
   \end{minipage}    
   \begin{minipage}[b]{.245\textwidth}
        \includegraphics[width=1 \textwidth,trim={0cm 9cm 0cm 0cm},clip]{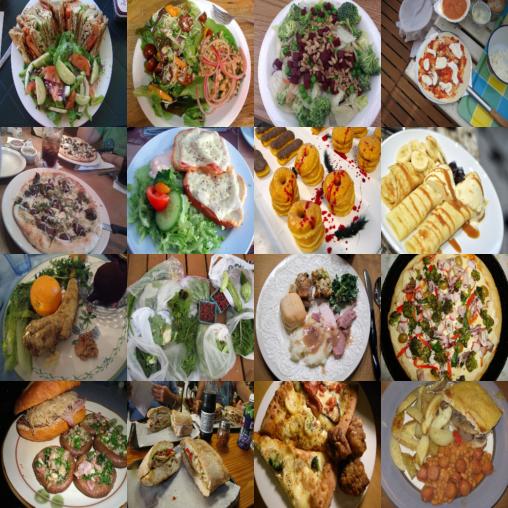}
   \end{minipage}
    \begin{minipage}[b]{.245\textwidth}
        \includegraphics[width=1 \textwidth,trim={0cm 9cm 0cm 0cm},clip]{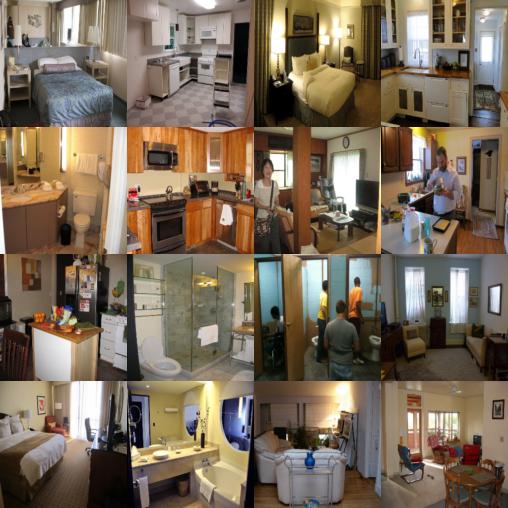}
   \end{minipage}  
  \caption{Blocks of the $8$ 
  most activating images for $4$ neurons of our network trained on ImageNet (top row) and COCO (bottom row). The counting neurons are sensitive to semantically similar images.  Interestingly, dominant concepts in each dataset, \eg, dogs in ImageNet and persons playing baseball in COCO, emerge in our counting vector.}
  \label{fig:activation}
\vspace{-.1in}
\end{figure*}

\subsubsection{Qualitative Analysis}

\noindent\textbf{Activating/Ignored images.} In Fig \ref{fig:high_low_norm}, we show blocks of 16 images ranked based on the magnitude of the counting vector. We observe that images with the lowest feature norms are textures without any high-level visual primitives. In contrast, images with the highest feature response mostly contain multiple object instances or a large object. For this experiment we use the validation or the test set of the dataset that the network has been trained on, so the network has not seen these images during training.

\noindent\textbf{Nearest neighbor search.} \hamed{To qualitatively evaluate our learned representation, for some validation images, we visualize their nearest neighbors in the training set in Fig.~\ref{fig:NN}.} Given a query image, the retrieval is obtained as a ranking of the Euclidean distance between the counting vector of the query image and the counting vector of images in the dataset. Smaller values indicate higher affinity. Fig.~\ref{fig:NN} shows that the retrieved results share a similar scene outline and are semantically related to the query images. Note that we perform retrieval in the counting space, which is the last layer of our network. This is different from the analogous experiment in \pf{}{the previous method }~\cite{krahenbuhl2015data} which performs the retrieval in the intermediate layers.  This result can be seen as an evidence that our initial hypothesis, that the counting vectors capture high level visual primitives, was true. 

\noindent\textbf{Neuron activations.} To visualize what each single counting neuron (\ie, feature element) has learned, we rank images not seen during training based on the magnitude of their neuron responses. We do this experiment on the validation set of ImageNet and the test set of COCO. In Fig.~\ref{fig:activation}, we show the top $8$ 
most activating images for $4$ neurons out of $30$ active ones on ImageNet and out of $44$ active ones on COCO. We observe that these neurons seem to cluster images that share the same scene layout and general content.

\section{Conclusions}

We have presented a novel representation learning method that does not rely on annotated data. We used counting as a pretext task, which we formalized as a constraint that relates the ``counted'' visual primitives in tiles of an image to those counted in its downsampled version. This constraint was used to train a neural network with a contrastive loss. Our experiments show that the learned features count non-trivial semantic content, qualitatively cluster images with similar scene outline, and outperform previous state of the art methods on transfer learning benchmarks. Our framework can be further extended to other tasks and transformations in addition to being combined with partially labeled data in a semi-supervised learning method.

\textbf{Acknowledgements.}
We thank Attila Szab\'{o} for insightful discussions about unsupervised learning and relations based on equivariance.
Paolo Favaro acknowledges support from the Swiss National Science Foundation on project 200021\_149227. Hamed Pirsiavash acknowledges support from GE Global Research.


{\small
\bibliographystyle{ieee}
\bibliography{counting}
}

\end{document}